\pdfoutput=1

\documentclass[11pt]{article}

\usepackage{acl}

\usepackage{times}
\usepackage{latexsym}

\usepackage[T1]{fontenc}

\usepackage[utf8]{inputenc}

\usepackage{microtype}

\usepackage{tabularx}
\usepackage{graphicx}
\usepackage{multirow}
\usepackage{booktabs}
\usepackage{enumitem}
\usepackage{multirow}
\usepackage{xspace}
\newcommand{\F}{F$_1$\xspace}
\usepackage{caption}
\usepackage{amsmath}

\newlist{compactitem}{itemize}{3}
\setlist[compactitem]{topsep=0pt,partopsep=0pt,itemsep=0pt,parsep=0pt}
\setlist[compactitem,1]{label=\textbullet}
\setlist[compactitem,2]{label=---}
\setlist[compactitem,3]{label=*}

\newlist{compactdesc}{description}{3}
\setlist[compactdesc]{topsep=0pt,partopsep=0pt,itemsep=0pt,parsep=0pt}

\newlist{compactenum}{enumerate}{3}
\setlist[compactenum]{topsep=0pt,partopsep=0pt,itemsep=0pt,parsep=0pt}
\setlist[compactenum,1]{label=(\arabic*)}
\setlist[compactenum,2]{label=\alph*}
\setlist[compactenum,3]{label=\roman*}

\usepackage{annotates}

\title{Entity-based Claim Representation Improves\\ Fact-Checking of Medical Content in Tweets}

\author{Amelie W\"uhrl \and Roman Klinger \\
  Institut f{\"u}r Maschinelle Sprachverarbeitung, University of Stuttgart, Germany \\
  \texttt{\{amelie.wuehrl,roman.klinger\}@ims.uni-stuttgart.de}\\
}

\begin{document}
\maketitle
\begin{abstract}
  False medical information on social media poses harm to people's
  health. While the need for biomedical fact-checking has been
  recognized in recent years, user-generated medical content has
  received comparably little attention. At the same time, models for
  other text genres might not be reusable, because the claims they
  have been trained with are substantially different. For instance,
  claims in the \textsc{SciFact} dataset are short and focused:
  ``\textit{Side effects associated with antidepressants increases
    risk of stroke}''. In contrast, social media holds
  naturally-occurring claims, often embedded in additional context:
  ``\textit{`If you take antidepressants like SSRIs, you could be at
    risk of a condition called serotonin syndrome' Serotonin syndrome
    nearly killed me in 2010. Had symptoms of stroke and seizure.}''
  This showcases the mismatch between real-world medical claims and
  the input that existing fact-checking systems expect. To make
  user-generated content checkable by existing models, we propose to
  reformulate the social-media input in such a way that the resulting
  claim mimics the claim characteristics in established datasets. To
  accomplish this, our method condenses the claim with the help of
  relational entity information and either compiles the claim out of
  an entity-relation-entity triple or extracts the shortest phrase
  that contains these elements. We show that the reformulated input
  improves the performance of various fact-checking models as opposed
  to checking the tweet text in its entirety.
\end{abstract}

\section{Introduction}
People use social media platforms like Twitter to discuss medical
issues. This can expose them to false health-related information and
poses immediate harm to people's well-being \citep{suarez-lledo_2021}.
While the necessity for fact-checking biomedical or scientific
information has been recognized and addressed in recent years,
naturally occurring arguments and claims as they are shared by social media users
have received less attention.

\begin{table}
\small
\setlength{\tabcolsep}{5pt} 
\begin{tabularx}{\columnwidth}{lp{12mm}X}
\toprule
Id & Source & Claim \\
\cmidrule(lr){1-1} \cmidrule(lr){2-2} \cmidrule(lr){3-3}
1 & \textsc{SciFact} & A mutation in HNF4A leads to an increased risk of diabetes by the age of 14 years.\\
2 & PubHealth & Scientists find clues to why binge-drinking causes binge-eating.\\
3 & Zuo et al. (\citeyear{zuo-etal-2020}) & Scientists discover gene mutation involved in paraplegia and epilepsy\\
4 & COVID-Fact & Baricitinib restrains the immune dysregulation in covid-19 patients\\
5 & HealthVer &Frequent touching of contaminated surfaces in public areas is therefore a potential route of SARS-CoV-2 transmission. \\
6 & Co\textsc{Vert} & So, they die from lung failure caused by extreme pneumonia or heart failure from sludgy blood but the root cause is \#COVID19 (which can be confirmed post-mortem) so the death is counted as due to the \#coronavirus \& NOT due to natural causes of pneumonia or heart attack\\
\bottomrule
\end{tabularx}
\caption{Claims from different fact-checking datasets.}
\label{table:claim-examples}
\end{table}

Unfortunately, systems trained on datasets from other domains might
not be reusable: The datasets that underly existing pretrained models
work with atomic, edited or summarized claims \citep[e.g., from
datasets like \textsc{SciFact},][]{wadden-et-al-2020}, cover claims
that have been selected to be well-formed
\citep[COVID-Fact,][]{saakyan-et-al-2021}, or contain editorial
content such as news headlines \citep{zuo-etal-2020}. Examples 1--5 in
Table~\ref{table:claim-examples} convey complex biomedical processes,
they are relatively short and coherently worded. In addition, they
make statements covering only one claim or fact. On the other hand,
medical statements as they organically occur for example on Twitter
are complex, wordy, imprecise and often ambiguous (Example~6). This
makes them substantially different to the claims in established
fact-checking datasets for the medical domain.
To address the limitations of using only well-formed claims, \citet{sarrouti-et-al-2021} propose a custom dataset and fact-checking model. Their analysis indicates that naturally occurring claims contain multiple,
inter-related facts compared to claims in other fact-verification
datasets. Along with \citet{zuo-et-al_2022}, they show that real-world
medical claims in user-generated and news content are more complex and
longer. In addition, \citet{kim-et-al-2021} show
that fact-checking systems do not transfer robustly to colloquial
claims.

This mismatch motivates extracting a check-worthy main claim from
user-generated content before continuing with fact-checking. This claim detection task, which is also a central task in argument mining, can be addressed as a sequence labeling problem \citep[i.a.]{zuo-et-al_2022}. While this
approach requires dedicated annotated data, we propose an alternative
that requires an entity annotation and relation detection system --
something that has been developed for various purposes across domains
\citep[i.a.]{yepes-macKinlay_2016, giorgi-bader_2018,
  scepanovic-et-at_2020, lamurias-et-al_2019,doan-et-al_2019,
  akkasi-moens_2021}. We hypothesize that the main information relevant to a claim is
encoded in entities and their relations, because they convey the key semantic information within a statement and describe how they interact with each other. For our approach we propose to use
that information to either find the claim token sequence or to
generate a sentence representation based on entity and relation
classes. Our results show that entity-based claim extraction supports
fact-checking for user-generated content, effectively making it more
accessible to MultiVerS \citep{wadden-et-al-2022}, an
architecture recently suggested for scientific claim verification.

\section{Related Work}
\subsection{Biomedical \& Scientific Fact-Checking}
The task of fact-checking is to determine the truthfulness of a claim
\citep{thorne-vlachos-2018}. This has been addressed for various
domains (\citet{guo-et-al_2021} provide a comprehensive review). For the general domain, some work has explored judging the
truthfulness of claims based on its linguistic features
\citep{rashkin-etal-2017} or using the knowledge stored in language
models as evidence \citep{lee-etal-2020}. Fact-checking for biomedical
and scientific content typically leverages external evidence sources.
In biomedicine this is vital as novel research that might change or overturn an existing view on a medical claim can only be taken into account if we tap into up to date, external evidence. In other fact-checking contexts (e.g., in a political context), this requirement is not as strong since the veracity of a statement made at a particular point in time is relatively stable. 
In the biomedical context, given a claim, fact-checking is typically modeled as a two-step process: evidence
retrieval (on document and/or sentence-level) and predicting a
verdict. This verdict either determines the veracity of the claim or
indicates if the evidence supports or refutes the claim.  We can group
existing approaches by the genre of text from which claims and
evidence stem.  \citet{wadden-lo-2021} formalize scientific claim
verification in the \textsc{Sciver} shared task, in which evidence and
claims both originate from expert-written
text. \citet{pradeep-et-al-2021} approach this task with a pipeline
model, while \citet{li-etal-2021, zhang-etal-2021} propose modeling one
or multiple subtasks in a multi-task learning setup.  Recently,
\citet{wadden-et-al-2022} showed that providing more context, i.e., by
representing the claim, full evidence abstract and title in a single
encoding, is beneficial for inferring a final verdict.

Moving away from expert-written text, \citet{kotonya-toni-2020}
explore verdict prediction for public health claims and use
fact-checking and news articles as
evidence. \citet{hossain-et.al-2020} classify a tweet into predefined
categories of known misconceptions about COVID-19.
\citet{mohr-wuehrl-klinger-2022} automatically verify tweets with
COVID-19-related claims with the help of excerpts from online sources.
Finally, some studies explore settings in which the claim and evidence
texts originate from different genres.  \citet{zuo-etal-2020}
investigate retrieving scientific evidence for biomedical claims in
news texts. \citet{sarrouti-et-al-2021} check user-generated, online
claims against scientific articles and \citet{saakyan-et-al-2021}
explore this task for COVID-19-related claims from Reddit.

\subsection{Datasets \& Their Claim Characteristics}
\label{related:datasets}
Various datasets have been proposed to facilitate scientific and
medical fact-checking.  One common characteristic lies in the claims
contained in these datasets: they are typically well-formed and
sometimes synthetic. This attribute presents a
misalignment with the type of data as it occurs on social media.

In \textsc{SciFact} \citep{wadden-et-al-2020} claims are
synthetic. They are atomic summaries of claims within scientific
articles. As evidence, the dataset provides abstracts from scientific
literature as well as sentence-level rationales for the claims within
those abstracts.  PubHealth \cite{kotonya-toni-2020} and the dataset
released by \citet{zuo-etal-2020} include claims from editorial
content. \citet{kotonya-toni-2020} provide claims and evidence texts
from health-related news and fact-checking articles while
\citet{zuo-etal-2020} identify the headlines of health news articles
as claims and provide the scientific papers referenced in the news article as evidence. While this genre of claims and content is
targeted towards non-experts, it undergoes journalistic editing and
can therefore not be characterized as occurring naturally.

We are aware of three datasets that cover user-generated claims, all
with a focus on COVID-19.  \textit{COVID-Fact}
\citep{saakyan-et-al-2021} contains medical claims shared on a
COVID-19-specific Sub-Reddit.  They use the scientific articles that
the users reference as evidence documents. The claims have been
filtered to retain only well-formed statements.
\citet{sarrouti-et-al-2021} contribute the \textit{HealthVer} corpus
of real-world statements from online users. To find relevant claims,
they query a search engine with COVID-19 questions and use the
resulting texts as claims. The provided evidence consists of abstracts
from scientific articles.  Similar, but exclusively focused on
COVID-19 information on Twitter, \textit{Co\textsc{Vert}}
\citep{mohr-wuehrl-klinger-2022} provides fact-checked tweets along
with evidence texts from online resources.  To the best of our
knowledge, only \textit{HealthVer} and \textit{Co\textsc{Vert}} cover naturally
occurring medical claims from a broad audience.

\subsection{Detecting, Extracting \& Generating Claims}
The task of claim detection is relevant to the field of fact-checking
as well as the area of argument mining.  From an argument mining
perspective, claim detection requires identifying the claim as the
core component within the argument structure
\citep{daxenberger-etal-2017}. While mainly rooted in the political
domain and social sciences \citep[i.a.]{lawrence-reed-2019, vecchi-etal-2021}, some work has explored
claim detection in scientific
text. \citet[i.a.]{achakulvisut-et-al_2019,
  mayer-et-al_2020,li-et-al-2021} extract claims from clinical and
biomedical articles, \citet{wuhrl-klinger-2021} classify tweets that
contain medical claims.

At the same time, detecting a checkable and check-worthy claim is
considered the first task within a fact-checking pipeline
\citep{guo-et-al_2021}. The task of claim-check-worthiness detection
is to determine if a given claim should be fact-checked. Typically
this is framed as a document, sentence or claim-level classification
or ranking task: \citet[i.a.]{gencheva-etal-2017, jaradat-etal-2018,
  wright-augenstein-2020} study this task for general domain claims,
in the \textit{CLEF-CheckThat!} shared task
\citep{clef-checkthat_2022} participants are tasked to identify tweets
that contain check-worthy claims about COVID-19. To the best of our
knowledge, \citet{zuo-et-al_2022} are the first to explore this on the
token level by extracting check-worthy claim sequences from
health-related news texts.  This shows that identifying biomedical
claim sequences in longer documents for the purpose of fact-checking
is understudied. The focus in fact-checking datasets and shared tasks
(e.g., \textsc{Fever} \citep{thorne-etal-2018} or \textsc{SciVER}
\citep{wadden-lo-2021}) is typically to infer the relationship between
a claim-evidence pair or on retrieving evidence for a given claim.

While in the studies described above the original phrasing of a
document or claim is kept intact, some work has proposed extracting
relevant semantic information to reconstruct the content that is being
conveyed. Recently, \citet{magnusson-friedman-2021} show that
fine-grained biomedical information within scientific text can be
extracted into a knowledge graph to model claims. Related to our work is 
\citet{yuan-yu-2019} who extract triplets from health-related news
headlines to capture medical claims. Their focus is on classifying the triples
as claim or non-claim which leaves fact-checking for future work. Our objective is to extract a concise claim representation and to explore its impact on fact-checking.

Moving even further away from the original text, \citet{wright-et-al-2022} suggest generating claims from
scientific text to address the data bottleneck for the downstream
fact-checking task. They report comparable performances for models
trained on automatically generated claims compared to a model trained
on the manually labeled \textsc{Scifact} claims. Their work is related
to \citet{pan-et-al-2021} who generate claims to facilitate zero-shot
fact verification for the general domain.

\section{Methods}
\label{methods}
With this work we investigate if knowledge about biomedical entities
allows us to extract a concise claim representation from
user-generated text that enables fact-checking systems to predict a verdict.  To explore this, we suggest two
methods to extract and construct entity-based, claim-like
statements. We assume we have a sequence of tokens
$\textbf{t} = (t_1, \ldots, t_n)$.
In addition, we have a set of $m$ annotations 
\begin{equation*}
A = \big\{(\,e^{\mathbf{a}_1}_{\textrm{subj}}, r^{\mathbf{a}_1}, e^{\mathbf{a}_1}_{\textrm{obj}}), \ldots, (\,e^{\mathbf{a}_m}_{\textrm{subj}}, r^{\mathbf{a}_m}, e^{\mathbf{a}_m}_{\textrm{obj}}){\big\rbrace},
\end{equation*}
which encode entity and relation information, respectively $e$ and $r$.
The entities are located within the token sequence $\textbf{t}$ and identified by their character-level onset $k$ and offset $\ell$ such that $e~=~(k, \ell)$, with $1  \leq k, \ell  \leq n$. The relation $r$ is a string representing the relation type (e.g., ``cause of'').

\begin{figure}[tb]
\centering
\includegraphics[scale=.8]{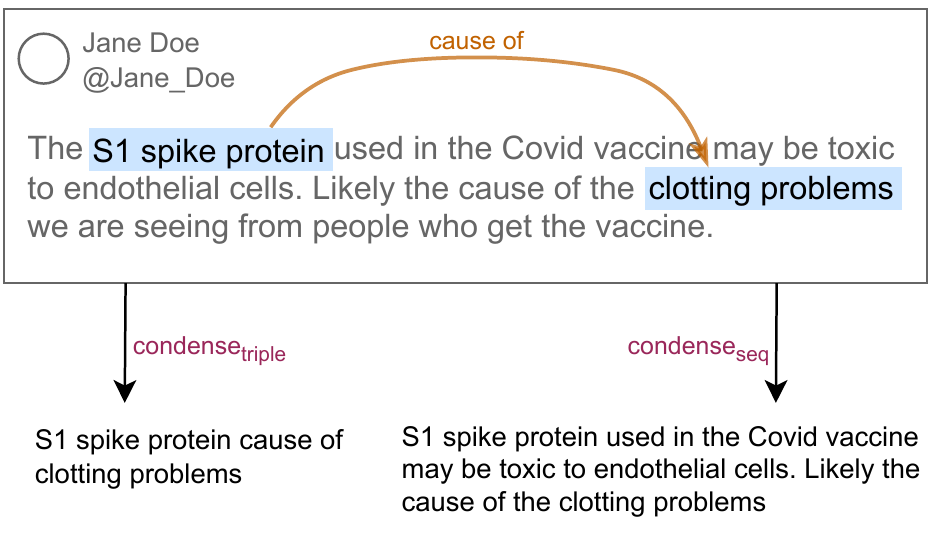}
\caption{Presented with an input document that has entity and relation
  labels, $\textrm{condense}_{\textrm{triple}}$ and
  $\textrm{condense}_{\textrm{seq}}$ extract two concise
  claims.}
\label{fig:condensation-example}
\end{figure}

Building on data of this type, we suggest two claim condensation
methods of the form 
\begin{equation*}
\textrm{condense}(\textbf{t},\textbf{a}) \rightarrow c
\end{equation*}
that transform the sequence $\mathbf{t}$ along with its annotation $a$ into a claim-like token sequence $c$. We propose two variants:

\paragraph{Representing Claims as Triples.}
\label{methods:claim-triple}
We reduce the claim to what we hypothesize to be its core components: two  medical entities and the relation between them. We hypothesize that the entities express the most relevant information with regard to the claim. 

In this representation the claim is a concatenation $\circ$ of the
subject entity tokens, the name of the relation $r$ and the object
entity tokens:
\begin{equation*}
\textrm{condense}_{\textrm{triple}}(\textbf{t}, \textbf{a}) = \mathbf{t}_{e^{\mathbf{a}}_{\textrm{subj}_{k}}: e^{\mathbf{a}}_{\textrm{subj}_\ell}} \circ r^{\mathbf{a}} \circ \mathbf{t}_{e^{\mathbf{a}}_{\textrm{obj}_{k}}: e^{\mathbf{a}}_{\textrm{obj}_\ell}}
\end{equation*}
This approach ignores tokens that are not part of the relation or entity annotation.

\paragraph{Extracting Claim Sequences.}
\label{methods:claim-seq}
Alternatively, we extract a subsequence from the original text. For
each annotation $\mathbf{a}$ in $\textbf{t}$, we apply
\begin{equation*}
\textrm{condense}_{\textrm{seq}}(\textbf{t}, \textbf{a}) = \mathbf{t}_{e^{\mathbf{a}}_{\textrm{subj}_{k}}}\cdots\circ\cdots{}\mathbf{t}_{e^{\mathbf{a}}_{\textrm{obj}_\ell}}.
\end{equation*}
This retains the way the author of the original text chose to express
the relation, including all tokens that are mentioned between the
entities. Commonly, this also involves words that indicate the
relation class, but we do not ensure
that. Figure~\ref{fig:condensation-example} shows examples of both
condensation methods. The example is taken from
\citet{mohr-wuehrl-klinger-2022}.

\section{Experiments}
\label{experiments}
We investigate whether we can reduce the complexity of user-generated
claims in order to make the information that they convey accessible to
pretrained ``off-the-shelf'' fact-checking models and circumvent the
necessity of custom training data and specialized models. We
specifically explore the use of entity information to formulate a
condensed version of a claim (see Section~\ref{methods}).  More
concretely, we compare how the claim representation impacts the
performance of a fact-checking model.

\subsection{Experimental Setting}

\subsubsection{Data}
\label{ex:data}
To test our claim condensation methods as outlined in Section
\ref{methods}, we assume the availability of entity and relation
information. This is not an unrealistic assumption: Entity and
relation extraction systems exist \citep[i.a.]{yepes-macKinlay_2016,
  giorgi-bader_2018, scepanovic-et-at_2020,
  lamurias-et-al_2019,doan-et-al_2019, akkasi-moens_2021}.  For our
experiments, we build on top of data that has such annotations to
focus the evaluation on the extraction method instead of evaluating
the quality of a NER/RE system. To the best of our knowledge, the only
dataset that provides both fact checking as well as entity and
relation information is the Co\textsc{Vert} corpus
\citep{mohr-wuehrl-klinger-2022}. The dataset consists of fact-checked
medical claims in tweets about COVID-19 and includes evidence texts
that the annotators provided to substantiate their verdicts
(\textsc{support}, \textsc{refute}, \textsc{not enough
  information}). Importantly, the dataset also contains the span and
type of medical entities and type of relations for each Twitter post.
The entity classes cover \textit{Medical Condition},
\textit{Treatment}, \textit{Symptom/Side-effect} and
\textit{Other}. Each tweet is also labeled with causative relations
\textit{(not\_)cause\_of} and \textit{causative\_agent\_of} between a
subject and an object entity. We use these annotations to formulate
the condensed claims.

Co\textsc{Vert} includes 300 tweets with a total of 722 entities and 300
relations.  In instances where multiple objects have been annotated
for an entity, we choose the triple which appears first in the
document under the assumption that the first claim is the main claim
of the statement. For short texts, such as tweets, we hypothesize that people will mention their central, main claim at the beginning of their statement. Additionally, this emulates the atomic nature of
claims in \textsc{Scifact}. Co\textsc{Vert} is crowd-annotated and provides
three evidence texts per claim. From those, we choose the first
snippet that is in line with the majority fact-checking verdict as the
gold evidence. While \textsc{Scifact} assigns the \textsc{Not enough
  info (NEI)} label if a given abstract does not provide enough
information to come to a verdict, in Co\textsc{Vert} a tweet is labeled as
\textsc{NEI} if annotators were not able to find any evidence or if
there was no majority w.r.t. the verdict. We therefore drop the 36
tweets labeled \textsc{NEI} for our experiments, as there is no
agreement w.r.t. the verdict class or no available evidence.  This
leaves us with 264 extracted claims (198 \textsc{Support}, 66
\textsc{Refutes}).

\subsubsection{Fact-checking Models}
\label{methods:fc-model}

We use the MultiVerS architecture which has recently been suggested
for evidence-based scientific fact verification
\citep{wadden-et-al-2022}. At the time of writing, this approach 
ranks first for the shared task \textsc{Sciver}.%
\footnote{\url{https://leaderboard.allenai.org/scifact/submissions/public}}
It takes as input a claim-evidence pair and represents both in a
single encoding to predict a fact-checking label and identify
rationales with the evidence. Claim, title and evidence abstract
sentences are concatenated using separator tokens and assigned global
attention during training. The model subsequently uses a classifier
over the separator token that identifies the claim to predict the
fact-checking verdict and an additional classification head over the
separator tokens between the evidence sentences.

Based on this architecture, \citet{wadden-et-al-2022} provide various
models.\footnote{We use their code base
  \url{https://github.com/dwadden/multivers} and the provided model
  checkpoints from there.} \textit{fever} is trained on the
\textsc{Fever} dataset for general domain
fact-checking. \textit{fever\_sci} is trained on a combination of
\textsc{Fever} data and weakly-labeled biomedical fact-checking
data. The other models build on top of \textit{fever\_sci} and are
subsequently fine-tuned on gold-labeled, in-domain data for verdict
prediction and rationale selection using \textit{scifact},
\textit{covidfact} and \textit{healthver}.

In order to test the impact of the claim representations, we do not adapt the fact-checking model, but alter the input claims.

\subsubsection{Baseline: Predicting Claim Sequences}
\label{ex:baseline}
To provide a baseline and gauge the impact of entity-based claim
representation as opposed to predicting a claim sequence without relying
on entities, we compare to the model by \citet{zuo-et-al_2022}. They
train a Bi-\textsc{Lstm-Crf} sequence labeling model to detect
check-worthy claims in medical news articles. Such articles are
similar to tweets in that they are also non-expert-written text
conveying medical information.  Using their code base and provided
training
data\footnote{\url{https://github.com/chzuo/jdsa_cross_genre_validation}},
we recreate their best performing model which encodes the input with a
combination of Bio\textsc{Bert} and \textsc{Flair}
embeddings.\footnote{\citet{zuo-et-al_2022} use the position of
  hyperlinks to a source publication within the news articles as
  additional input to their model. However, they report that the
  performance gains using this information is not statistically
  significant. As the Co\textsc{Vert} data does not contain this type of
  information, we do not include it when recreating their model.}
We use the resulting model to predict claim sequences in
the Co\textsc{vert} tweets\footnote{We make predictions for 264 Co\textsc{Vert}  tweets not labeled as \textsc{NEI} (see Sec. \ref{ex:data}).}. For tweets where the model predicted more than one
claim sequence in a tweet we use the prediction with the highest
confidence score. Note that for 6 tweets the model does not predict
any claim. This leaves us with 258 claims.

\subsubsection{Evaluation}
We evaluate the claim condensation techniques on the downstream task of predicting a fact-checking verdict for a claim-evidence pair.
Following \citet{wadden-et-al-2022} we report the \textit{Label-Only} \F on abstract level from the
\textsc{SciFact} task\footnote{We use their evaluation script:
  \url{https://github.com/allenai/scifact-evaluator}}. It measures the \F-score of the model
for predicting the correct fact-checking verdict given a claim and evidence
candidate. A true positive is therefore a claim-evidence pair with a
correctly predicted verdict.

 \begin{table*}[tb]
    \centering \small
    \setlength{\tabcolsep}{5pt} 
    \begin{tabularx}{\textwidth}{lrrrrrrrrrrrrrrr}
    \toprule
        & \multicolumn{15}{c}{Claim Representation}\\
        \cmidrule(l){2-16}
        & \multicolumn{3}{c}{full tweets} & \multicolumn{4}{c}{\citet{zuo-et-al_2022}} & \multicolumn{4}{c}{$\textrm{condense}_{\textrm{triple}}$} & \multicolumn{4}{c}{$\textrm{condense}_{\textrm{seq}}$} \\
        \cmidrule(lr){2-4} \cmidrule(lr){5-8} \cmidrule(lr){9-12} \cmidrule(l){13-16}
        model& P & R & \F & P & R & \F & $\Delta$ & P & R & \F & $\Delta$ & P & R & \F & $\Delta$\\
         
         \cmidrule(r){1-1}
      \cmidrule(lr){2-2}\cmidrule(lr){3-3}\cmidrule(lr){4-4}
      \cmidrule(lr){5-5}\cmidrule(lr){6-6}\cmidrule(lr){7-7}\cmidrule(lr){8-8}
      \cmidrule(lr){9-9}\cmidrule(lr){10-10}\cmidrule(lr){11-11}\cmidrule(lr){12-12}
      \cmidrule(lr){13-13}\cmidrule(lr){14-14}\cmidrule(lr){15-15}\cmidrule(l){16-16}
      
    		fever & 
    		0.0 & 0.0 & 0.0 
    		& 75.0 & 1.2 & 2.3 & $+$2.3
    		& 81.8 & 3.4 & \textbf{6.5} & $+$6.5
    		& 83.3 & 1.9 & 3.7 & $+$3.7 \\
    		
    		fever\_sci & 
    		91.7 & 4.2 & 8.0 
    		& 100 & 10.1 & 18.3 & $+$10.3 
    		& 89.8 & 20.1 & \textbf{32.8} & $+$24.8
    		& 87.2 & 15.5 & 26.4 & $+$18.4 \\

    		scifact & 
    		100 & 0.4 & 0.8 
    		& 100 & 2.7 & 5.3 & $+$4.5
		& 86.4 & 7.2 & 13.3 & $+$12.5
    		& 90.9 & 7.6 & \textbf{14.0} & $+$13.2 \\

    		covidfact & 
    		30.8 & 4.5 & 7.9 
		& 48.6 & 14.0 & 21.7 & $+$13.8
		& 65.0 & 30.3 & \textbf{41.3} & $+$33.4
    		& 55.6 & 28.4 & 37.6 & $+$29.7\\

    		healthver & 
    		82.8 & 31.1 & 45.2 
		& 86.9 & 33.3 & 48.2  & $+$3.0
		& 79.7 & 41.7 & 54.7  & $+$9.5
    		& 85.9 & 48.5 & \textbf{62.0} & $+$16.8 \\
    		
    		\cmidrule(lr){2-4} \cmidrule(lr){5-8} \cmidrule(lr){9-12} \cmidrule(lr){13-16}
    		average & 61.1 & 8.0 & 12.4 
    		& 82.1 & 12.3 & 19.2 & $+$6.8
		& 80.5 & 20.5 & \textbf{29.7} & $+$17.3
		& 80.6 & 20.4 & 28.7 & $+$16.3	
    		\\

    \bottomrule
    \end{tabularx}
    \caption{Fact-checking performance of MultiVerS-based models
      (\textit{fever, fever\_sci, scifact, covidfact, healthver}) on
      Co\textsc{Vert} data. As the claim input, we present the model with the
      full tweets, a sequence predicted to contain the claim
      \citep{zuo-et-al_2022}, and the claims that we obtain from our
      entity and relation-based extraction methods
      $\textrm{condense}_{\textrm{triple}}$ and
      $\textrm{condense}_{\textrm{seq}}$. We report precision, recall
      and \F. For each model, $\Delta$ captures the difference in \F
      between the full tweet as input and the claims obtained from the
      respective claim detection or extraction methods. The last row denotes the average across all models. The best
      performance for each model is printed in bold face.}
    \label{tab:longchecker-versions-on-covert}
\end{table*}

\subsection{Results}
We report results for four approaches to represent
the claim. Our baselines are:
\begin{compactdesc}
\item [full] Full text of the tweet which contains a claim.
\item [Zuo et al. (2022)] A sequence predicted by a claim detection model, not informed by entity or relation knowledge.
\end{compactdesc}
The methods that we propose are:
\begin{compactdesc}
\item [$\textrm{condense}_{\textrm{triple}}$] Claim represented by an entity--relation triple.
\item [$\textrm{condense}_{\textrm{seq}}$] Shortest token
  sequence which contains all entities.
\end{compactdesc}

Table~\ref{tab:longchecker-versions-on-covert} reports the
results. The columns indicate which type of claim the models receive
as input. For each claim type and model we report precision, recall and \F as well as the difference $\Delta$ in \F to the prediction performance for the \textbf{full} tweet.
The table rows denote which model is used for prediction. The models (\textit{fever, fever\_sci, scifact, covidfact, healthver}) are based on the
MultiVerS architecture and vary w.r.t.\ the type of data they were
trained on.

Overall, we observe three major patterns from the results: (1)~All
models show limited performance when presented with the full
tweet. (2)~Delimiting the claim sequence always improves verdict
prediction. (3)~Representing the claim based on the entities and
relations is highly beneficial and leads to the most successful
predictions.  In the following, we discuss the results in more detail.

\paragraph{Fact-checking models struggle to predict verdicts for full
  tweets.}
In the first block of Table~\ref{tab:longchecker-versions-on-covert},
we see that the performance is generally low (avg.\ \F=12.4) when the
models are tasked to check the full tweet. The \textit{fever} model fails to predict fact-checking verdicts for this type of input. The \textit{healthver}
model is the most successful (\F=45.2), presumably because its
training data fits the Co\textsc{Vert} data best.

\paragraph{Delimiting the claim sequence is beneficial.}
Using the claim sequence prediction obtained with the
\citet{zuo-et-al_2022} model as claim input shows an improved
performance across all models (increases between 2.3 and 13.8pp in \F
compared to predictions for the full tweet). \textit{healthver}
remains the most successful model (48.2 \F). Notably, the
\textit{covidfact} model benefits most from the adapted input
($\Delta$ 13.8pp in \F).

\paragraph{Entity-based claim condensation improves verdict prediction.}
Across all models, one of the entity and relation-based claim
representations achieves the best results.
For three out of five models, $\textrm{condense}_{\textrm{triple}}$
claims facilitate the best prediction compared to other input
types. \textit{fever}, \textit{fever\_sci} and \textit{covidfact}
achieve \F-scores of 6.5, 32.8 and 41.3, respectively.
For the \textit{scifact} and \textit{healthver} model, using the
$\textrm{condense}_{\textrm{seq}}$ extracted claims leads to the most
reliable predictions: we observe 14.0 \F for \textit{scifact} and 62.0
\F for \textit{healthver}. \textit{healthver}'s prediction for the
$\textrm{condense}_{\textrm{seq}}$ claims is the most successful
across all models and settings.

Across the board, the \textit{covidfact} model benefits the most from
delimiting the claim sequence. Here, we observe increases in \F of
13.8, 33.4 and 29.7pp when comparing the performance on the full tweet
with that for a \citet{zuo-et-al_2022} claim,
$\textrm{condense}_{\textrm{triple}}$ and
$\textrm{condense}_{\textrm{seq}}$ claim, respectively.

The results show that both $\textrm{condense}$ methods improve the
performance of the fact-checking models. While the \F-scores and the
improvements ($\Delta$ values) vary across models, we observe the same
pattern across our experiments: providing a concise claim as input
leads to a more reliable verdict prediction. We also see that claims
from both $\textrm{condense}$ methods are more successfully checked
than the predicted claim sequence identified by the
\citet{zuo-et-al_2022} model. This shows that entities and relations
do capture the core information of a claim relatively well. It is
important to note that the $\textrm{condense}$ claims are constructed
using gold annotated entities and relations from Co\textsc{Vert},
while the predicted sequence is not. This needs to be taken into
account when comparing the results for those claim representations.

\begin{table*}[t]
  \centering \small
  \setlength{\tabcolsep}{5pt}
  \begin{tabularx}{\textwidth}{lXp{23mm}ccc}
    \toprule
    
    & \multicolumn{2}{c}{claim input} &\multicolumn{2}{c}{Pred.} & \\
    \cmidrule(lr){2-3} \cmidrule(lr){4-5}
    id & full tweet & $\textrm{cond}_{\textrm{seq}}$ & F & C & G\\
    \cmidrule(lr){1-1} \cmidrule(lr){2-2} \cmidrule(lr){3-3} \cmidrule(lr){4-4} \cmidrule(lr){5-5} \cmidrule(lr){6-6} 
    
    1a & Actually wearing masks causes bacterial pneumonia which people can die from NOT covid19. Most people do not know how to don/doff PPE properly. Follow the science Big Guy! & masks causes bacterial pneumonia 
     	& R & R & R \\

     	1b & Up to half of hospitalized COVID patients have elevated levels of antiphospholipid antibodies, or antibodies that cause blood clots to form. Patients with these antibodies are much more likely to have severe respiratory disease and kidney injury. \#COVID19 & elevated levels of antiphospholipid antibodies, or antibodies that cause blood clots to form 
     	&S &S & S\\
     	
     	\cmidrule(lr){1-1}\cmidrule(lr){2-6}
     	
		2a & ``It’s unclear if his death was related to the virus.” This is why we perform autopsies. There is a significant likelihood that \#COVID19 played a role in that it is known to affect endothelial cells \& has been shown to cause neurological symptoms including stroke. & COVID19 played a role in that it is known to affect endothelial cells \& has been shown to cause neurological symptoms 
		& N & S & S\\

     	2b &Are you aware that the vaccines could cause miscarriage? The real data regarding covid is that there are tiny numbers, percentage wise, of generally healthy people under the age of 60 that die from COVID or that get admitted into ICU. Are you worried about cancer too? & vaccines could cause miscarriage
     	& S& R & R \\
     	\cmidrule(lr){1-1} \cmidrule(lr){2-6}
     	
     	3a &The predominant symptoms of `long COVID’ are psychological in nature, with anxiety and depression being most common. But those of course are also exactly the conditions which have been caused in, literally, millions of people, especially young people, by the lockdowns. & long COVID’ are psychological in nature, with anxiety 
     	& S& N & S\\
     	
     	3b & Know the facts! There is no evidence that \#COVID19 \#vaccines cause \#infertility, says @username @username \& @username \#NIAW2021 \#InfertilityAwareness &COVID19 \#vaccines cause \#infertility
     	 &S & R & S \\
     	
     	\cmidrule(lr){1-1} \cmidrule(lr){2-6}
		
         4a & Covid is no joke, this is why we need the vaccine. We know that mRNA doesn't cause long term effects since it decomposes in your body within 1-2 hours. Please everyone, get vaccinated as soon as you can!& mRNA doesn't cause long term effects 
         & N&  N & S \\
         
         4b & I never said Covid-19 wasn’t a real coronavirus.  And deaths linked to Covid-19 are primarily caused directly from pneumonia, or flu-like symptoms.  The classifications for influenza and pneumonia reporting changed when Covid-19 appeared. Facts. & deaths linked to Covid-19 are primarily caused directly from pneumonia 
         & S &S & R\\  
           
    \bottomrule
    \end{tabularx}
    \caption{Example predictions for full tweets
      vs.\ $\textrm{condense}_{\textrm{seq}}$ claims. For each error category, we provide two examples (a and b). The predictions are
      made by the \textit{healthver} model. F: full
      tweet as input, C: condensed with method $\textrm{condense}_{\textrm{seq}}$, G: gold
      annotation. S: Supports, R: Refutes, N: Not enough
      information.}
    \label{tab:error-analysis}
\end{table*}

\section{Analysis and Discussion}
We aim to understand in which cases condensing the claim is helpful
and when it harms the performance. We therefore conduct an error
analysis where we compare the predictions of the best model
(\textit{healthver}) with the full tweet as input with predictions of
that model using the claims from the most successful condensation
method $\textrm{condense}_{\textrm{seq}}$. The examples mentioned in
the following section are displayed in Table
\ref{tab:error-analysis}. For the sake of brevity, we provide the
relevant evidence documents in the Appendix, Table
\ref{tab:covert-with-evidence}.

In total, there are 54 instances in which both claim inputs lead to a
correct label. In those instances, the tweet itself tends to be fairly
short (see Ex.~1a) or relatively well-formed (see Ex.~1b).

There are 74 instances in which the condensed claim sequence produces
a correctly predicted label while the check based on the full tweet
input does not lead to a correct result. For 66 out of 74, we observe
that the label flips from \textsc{NEI} to the correct label (see
Ex.~2a). This shows that the condensation can make the evidence more
accessible to the fact-checker. In addition, Ex.~2b shows how a
condensed claim is assigned a correct label, while the full tweet is
not. This might be the case because the claim is presented as a
question in the tweet.

In 28 cases condensing the claim leads to an incorrect prediction
while checking the full tweet leads to a correct output. In 20 cases,
condensing the claim changes the predicted label from the gold verdict
to \textsc{NEI} (see Ex.~3a). This indicates that condensation can
render evidence unusable.  In Example~3b the condensation actually
misrepresents the statement because it cuts of the phrase `no
evidence' before the claim. We recognize that this is a potential
pitfall of the claim extraction methods.

There are 108 instances where both claim types lead to incorrectly
predicted labels. In 90 out of 108 cases, both are labeled with
\textsc{NEI}. The evidence did not provide sufficient information to
check the claim.  Example~4a exemplifies that, to a certain degree,
the \textsc{NEI} label makes sense. The evidence (see
Table~\ref{tab:covert-with-evidence}) does not specifically mention
long-term consequences of mRNA (vaccines). To conclude that the claim
is supported by the evidence, we need to infer that long-term effects
are improbable, because the mRNA does not stay in the body or affect
the DNA. Similarly, in 4b, the evidence (see
Table~\ref{tab:covert-with-evidence}) requires reasoning, because
`pneunomia' and `flu-like symptoms' which the tweet claims are primary
causes of death in COVID-19 patients are not mentioned directly in the
evidence. In addition, the comparative statement in the evidence of
septic shock and multi-organ failure being the more prevalent causes
of death as opposed to respiratory failure might pose difficulties for
the model.

\section{Conclusion and Future Work}

Based on the substantial mismatch between the biomedical claims as
they are most typically expected by existing fact-checking models and
the nature of real-world, user-generated medical statements made on
Twitter, we propose to extract entity-based claim representations. We use the entities
as the core information relevant to the claim, and extract condensed
claims from tweets. When presented with the adapted claim input, the
fact-checking models we experiment with are able to verify the claims
more reliably as opposed to when they are tasked to infer a verdict
for a full tweet.

In this study, we focused the analysis on an existing dataset with a
comparably narrow focus. While we intuitively believe that the
findings also hold for other domains, this remains to be
proven. Therefore we propose that future work explores entity- and
relation-based claim extraction for other types of medical
relations. Co\textsc{Vert} focuses on causative claims which are by
design of the dataset explicitly mentioned in the tweet. Exploring if
claims about other types of relations can be extracted in a similar
manner is up to future research. Similarly, it is important to explore
how this method translates to statements with more than one
entity-relation-entity triple.

Another limitation of our analysis is its focus on one
fact-checking architecture. It is important to evaluate if the impact of claim condensation carries over to other claim checking
methods. A possible alternative to our approach (change the claim at
test time) could also be to adapt the system (adapt the claims at
training time). The degree of which the difference in genre and structure of
the evidence document might impact the models' performances is another
important perspective for future research.

Finally, we performed studies based on correct annotations of
entities. While this is a reasonable approach in a research
environment, it is important to explore the impact of error
propagation from a named entity recognizer to claim
condensation.

Apart from verdict prediction, entity-based claim representation could
also facilitate discovering suitable evidence for user-generated
medical content as entity knowledge has been shown to benefit evidence
retrieval as well \citep{hanselowski-et-al-2018}.

\section{Ethical Considerations}
Unreliable fact-checking evidence and verdicts potentially exacerbate
the spread of misinformation because they lend false credibility to
harmful health-related information. Therefore, it is greatly important
to carefully evaluate and analyze automatic fact-checking systems
before their predictions can be used reliably.

It is important to acknowledge that by extracting a claim sequence
from a broader statement, we might omit essential context. This could
impact the statement's meaning, its intended gravity or generally
misrepresent the claim that the author originally meant to convey. To
alleviate this and contextualize an automatically generated verdict,
it is important to design applications which are transparent with
respect to the input claims and prediction process.

\section*{Acknowledgements}
This research has been conducted as part of the FIBISS project which is funded by the German Research Council (DFG, project number: KL 2869/5-1). We thank the anonymous reviewers for their valuable feedback.

\bibliography{literature}

\begin{thebibliography}{42}
\expandafter\ifx\csname natexlab\endcsname\relax\def\natexlab#1{#1}\fi

\bibitem[{Achakulvisut et~al.(2019)Achakulvisut, Bhagavatula, Acuna, and
  K{\"{o}}rding}]{achakulvisut-et-al_2019}
Titipat Achakulvisut, Chandra Bhagavatula, Daniel~E. Acuna, and Konrad~P.
  K{\"{o}}rding. 2019.
\newblock \href {http://arxiv.org/abs/1907.00962} {Claim extraction in
  biomedical publications using deep discourse model and transfer learning}.
\newblock \emph{CoRR}, abs/1907.00962.

\bibitem[{Akkasi and Moens(2021)}]{akkasi-moens_2021}
Abbas Akkasi and Mari-Francine Moens. 2021.
\newblock \href {https://doi.org/https://doi.org/10.1016/j.jbi.2021.103820}
  {Causal relationship extraction from biomedical text using deep neural
  models: A comprehensive survey}.
\newblock \emph{Journal of Biomedical Informatics}, 119:103820.

\bibitem[{Daxenberger et~al.(2017)Daxenberger, Eger, Habernal, Stab, and
  Gurevych}]{daxenberger-etal-2017}
Johannes Daxenberger, Steffen Eger, Ivan Habernal, Christian Stab, and Iryna
  Gurevych. 2017.
\newblock \href {https://doi.org/10.18653/v1/D17-1218} {What is the essence of
  a claim? cross-domain claim identification}.
\newblock In \emph{Proceedings of the 2017 Conference on Empirical Methods in
  Natural Language Processing}, pages 2055--2066, Copenhagen, Denmark.
  Association for Computational Linguistics.

\bibitem[{Doan et~al.(2019)Doan, Yang, Tilak, Li, Zisook, and
  Torii}]{doan-et-al_2019}
Son Doan, Elly~W. Yang, Sameer~S. Tilak, Peter~W. Li, Daniel~S. Zisook, and
  Manabu Torii. 2019.
\newblock \href {https://doi.org/10.1186/s12911-019-0785-0} {Extracting
  health-related causality from twitter messages using natural language
  processing}.
\newblock \emph{{BMC} Medical Informatics and Decision Making}, 19(3):79.

\bibitem[{Gencheva et~al.(2017)Gencheva, Nakov, M{\`a}rquez,
  Barr{\'o}n-Cede{\~n}o, and Koychev}]{gencheva-etal-2017}
Pepa Gencheva, Preslav Nakov, Llu{\'\i}s M{\`a}rquez, Alberto
  Barr{\'o}n-Cede{\~n}o, and Ivan Koychev. 2017.
\newblock \href {https://doi.org/10.26615/978-954-452-049-6_037} {A
  context-aware approach for detecting worth-checking claims in political
  debates}.
\newblock In \emph{Proceedings of the International Conference Recent Advances
  in Natural Language Processing, {RANLP} 2017}, pages 267--276, Varna,
  Bulgaria. INCOMA Ltd.

\bibitem[{Giorgi and Bader(2018)}]{giorgi-bader_2018}
John~M Giorgi and Gary~D Bader. 2018.
\newblock \href {https://doi.org/10.1093/bioinformatics/bty449} {Transfer
  learning for biomedical named entity recognition with neural networks}.
\newblock \emph{Bioinformatics}, 34(23):4087--4094.

\bibitem[{Guo et~al.(2022)Guo, Schlichtkrull, and Vlachos}]{guo-et-al_2021}
Zhijiang Guo, Michael Schlichtkrull, and Andreas Vlachos. 2022.
\newblock \href {https://doi.org/10.1162/tacl_a_00454} {A survey on automated
  fact-checking}.
\newblock \emph{Transactions of the Association for Computational Linguistics},
  10:178--206.

\bibitem[{Hanselowski et~al.(2018)Hanselowski, Zhang, Li, Sorokin, Schiller,
  Schulz, and Gurevych}]{hanselowski-et-al-2018}
Andreas Hanselowski, Hao Zhang, Zile Li, Daniil Sorokin, Benjamin Schiller,
  Claudia Schulz, and Iryna Gurevych. 2018.
\newblock \href {https://doi.org/10.18653/v1/W18-5516} {{UKP}-athene:
  Multi-sentence textual entailment for claim verification}.
\newblock In \emph{Proceedings of the First Workshop on Fact Extraction and
  {VER}ification ({FEVER})}, pages 103--108, Brussels, Belgium. Association for
  Computational Linguistics.

\bibitem[{Hossain et~al.(2020)Hossain, Logan~IV, Ugarte, Matsubara, Young, and
  Singh}]{hossain-et.al-2020}
Tamanna Hossain, Robert~L. Logan~IV, Arjuna Ugarte, Yoshitomo Matsubara, Sean
  Young, and Sameer Singh. 2020.
\newblock \href {https://doi.org/10.18653/v1/2020.nlpcovid19-2.11}
  {{COVIDL}ies: Detecting {COVID}-19 misinformation on social media}.
\newblock In \emph{Proceedings of the 1st Workshop on {NLP} for {COVID}-19
  (Part 2) at {EMNLP} 2020}, Online. Association for Computational Linguistics.

\bibitem[{Jaradat et~al.(2018)Jaradat, Gencheva, Barr{\'o}n-Cede{\~n}o,
  M{\`a}rquez, and Nakov}]{jaradat-etal-2018}
Israa Jaradat, Pepa Gencheva, Alberto Barr{\'o}n-Cede{\~n}o, Llu{\'\i}s
  M{\`a}rquez, and Preslav Nakov. 2018.
\newblock \href {https://doi.org/10.18653/v1/N18-5006} {{C}laim{R}ank:
  Detecting check-worthy claims in {A}rabic and {E}nglish}.
\newblock In \emph{Proceedings of the 2018 Conference of the North {A}merican
  Chapter of the Association for Computational Linguistics: Demonstrations},
  pages 26--30, New Orleans, Louisiana. Association for Computational
  Linguistics.

\bibitem[{Kim et~al.(2021)Kim, Kim, Hong, and Kim}]{kim-et-al-2021}
Byeongchang Kim, Hyunwoo Kim, Seokhee Hong, and Gunhee Kim. 2021.
\newblock \href {https://doi.org/10.18653/v1/2021.naacl-main.121} {How robust
  are fact checking systems on colloquial claims?}
\newblock In \emph{Proceedings of the 2021 Conference of the North American
  Chapter of the Association for Computational Linguistics: Human Language
  Technologies}, pages 1535--1548, Online. Association for Computational
  Linguistics.

\bibitem[{Kotonya and Toni(2020)}]{kotonya-toni-2020}
Neema Kotonya and Francesca Toni. 2020.
\newblock \href {https://doi.org/10.18653/v1/2020.emnlp-main.623} {Explainable
  automated fact-checking for public health claims}.
\newblock In \emph{Proceedings of the 2020 Conference on Empirical Methods in
  Natural Language Processing (EMNLP)}, pages 7740--7754, Online. Association
  for Computational Linguistics.

\bibitem[{Lamurias et~al.(2019)Lamurias, Sousa, Clarke, and
  Couto}]{lamurias-et-al_2019}
Andre Lamurias, Diana Sousa, Luka~A. Clarke, and Francisco~M. Couto. 2019.
\newblock \href {https://doi.org/10.1186/s12859-018-2584-5} {{BO}-{LSTM}:
  classifying relations via long short-term memory networks along biomedical
  ontologies}.
\newblock \emph{{BMC} Bioinformatics}, 20(1):10.

\bibitem[{Lawrence and Reed(2019)}]{lawrence-reed-2019}
John Lawrence and Chris Reed. 2019.
\newblock \href {https://doi.org/10.1162/coli_a_00364} {Argument mining: A
  survey}.
\newblock \emph{Computational Linguistics}, 45(4):765--818.

\bibitem[{Lee et~al.(2020)Lee, Li, Wang, Yih, Ma, and Khabsa}]{lee-etal-2020}
Nayeon Lee, Belinda~Z. Li, Sinong Wang, Wen-tau Yih, Hao Ma, and Madian Khabsa.
  2020.
\newblock \href {https://doi.org/10.18653/v1/2020.fever-1.5} {Language models
  as fact checkers?}
\newblock In \emph{Proceedings of the Third Workshop on Fact Extraction and
  VERification (FEVER)}, pages 36--41, Online. Association for Computational
  Linguistics.

\bibitem[{Li et~al.(2021{\natexlab{a}})Li, Burns, and Peng}]{li-etal-2021}
Xiangci Li, Gully Burns, and Nanyun Peng. 2021{\natexlab{a}}.
\newblock \href {http://ceur-ws.org/Vol-2831/paper8.pdf} {A paragraph-level
  multi-task learning model for scientific fact-verification}.
\newblock In \emph{Proceedings of the Workshop on Scientific Document
  Understanding co-located with 35th {AAAI} Conference on Artificial
  Inteligence ({AAAI} 2021)}, Online.

\bibitem[{Li et~al.(2021{\natexlab{b}})Li, Burns, and Peng}]{li-et-al-2021}
Xiangci Li, Gully Burns, and Nanyun Peng. 2021{\natexlab{b}}.
\newblock \href {https://doi.org/10.18653/v1/2021.eacl-main.218} {Scientific
  discourse tagging for evidence extraction}.
\newblock In \emph{Proceedings of the 16th Conference of the European Chapter
  of the Association for Computational Linguistics: Main Volume}, pages
  2550--2562, Online. Association for Computational Linguistics.

\bibitem[{Magnusson and Friedman(2021)}]{magnusson-friedman-2021}
Ian Magnusson and Scott Friedman. 2021.
\newblock \href {https://doi.org/10.18653/v1/2021.emnlp-main.381} {Extracting
  fine-grained knowledge graphs of scientific claims: Dataset and
  transformer-based results}.
\newblock In \emph{Proceedings of the 2021 Conference on Empirical Methods in
  Natural Language Processing}, pages 4651--4658, Online and Punta Cana,
  Dominican Republic. Association for Computational Linguistics.

\bibitem[{Mayer et~al.(2020)Mayer, Cabrio, and Villata}]{mayer-et-al_2020}
Tobias Mayer, Elena Cabrio, and Serena Villata. 2020.
\newblock \href {https://hal.archives-ouvertes.fr/hal-02879293}
  {{Transformer-based Argument Mining for Healthcare Applications}}.
\newblock In \emph{{ECAI 2020 - 24th European Conference on Artificial
  Intelligence}}, Santiago de Compostela / Online, Spain.

\bibitem[{Mohr et~al.(2022)Mohr, Wührl, and
  Klinger}]{mohr-wuehrl-klinger-2022}
Isabelle Mohr, Amelie Wührl, and Roman Klinger. 2022.
\newblock \href {https://aclanthology.org/2022.lrec-1.26} {Covert: A corpus of
  fact-checked biomedical covid-19 tweets}.
\newblock In \emph{Proceedings of the Language Resources and Evaluation
  Conference}, pages 244--257, Marseille, France. European Language Resources
  Association.

\bibitem[{Nakov et~al.(2022)Nakov, Barr\'{o}n-Cede\~{n}o, Da~San~Martino, Alam,
  M\'{\i}guez, Caselli, Kutlu, Zaghouani, Li, Shaar, Mubarak, Nikolov, Kartal,
  and Beltr\'{a}n}]{clef-checkthat_2022}
Preslav Nakov, Alberto Barr\'{o}n-Cede\~{n}o, Giovanni Da~San~Martino, Firoj
  Alam, Rub\'{e}n M\'{\i}guez, Tommaso Caselli, Mucahid Kutlu, Wajdi Zaghouani,
  Chengkai Li, Shaden Shaar, Hamdy Mubarak, Alex Nikolov, Yavuz~Selim Kartal,
  and Javier Beltr\'{a}n. 2022.
\newblock \href {http://ceur-ws.org/Vol-2936/paper-28.pdf} {Overview of the
  {CLEF}-2022 {CheckThat}! lab task 1 on identifying relevant claims in
  tweets}.
\newblock In \emph{Working Notes of CLEF 2022---Conference and Labs of the
  Evaluation Forum}, CLEF~'2022, Bologna, Italy.

\bibitem[{Pan et~al.(2021)Pan, Chen, Xiong, Kan, and Wang}]{pan-et-al-2021}
Liangming Pan, Wenhu Chen, Wenhan Xiong, Min-Yen Kan, and William~Yang Wang.
  2021.
\newblock \href {https://doi.org/10.18653/v1/2021.acl-short.61} {Zero-shot fact
  verification by claim generation}.
\newblock In \emph{Proceedings of the 59th Annual Meeting of the Association
  for Computational Linguistics and the 11th International Joint Conference on
  Natural Language Processing (Volume 2: Short Papers)}, pages 476--483,
  Online. Association for Computational Linguistics.

\bibitem[{Pradeep et~al.(2021)Pradeep, Ma, Nogueira, and
  Lin}]{pradeep-et-al-2021}
Ronak Pradeep, Xueguang Ma, Rodrigo Nogueira, and Jimmy Lin. 2021.
\newblock \href {https://aclanthology.org/2021.louhi-1.11} {Scientific claim
  verification with {V}er{T}5erini}.
\newblock In \emph{Proceedings of the 12th International Workshop on Health
  Text Mining and Information Analysis}, pages 94--103, online. Association for
  Computational Linguistics.

\bibitem[{Rashkin et~al.(2017)Rashkin, Choi, Jang, Volkova, and
  Choi}]{rashkin-etal-2017}
Hannah Rashkin, Eunsol Choi, Jin~Yea Jang, Svitlana Volkova, and Yejin Choi.
  2017.
\newblock \href {https://doi.org/10.18653/v1/D17-1317} {Truth of varying
  shades: Analyzing language in fake news and political fact-checking}.
\newblock In \emph{Proceedings of the 2017 Conference on Empirical Methods in
  Natural Language Processing}, pages 2931--2937, Copenhagen, Denmark.
  Association for Computational Linguistics.

\bibitem[{Saakyan et~al.(2021)Saakyan, Chakrabarty, and
  Muresan}]{saakyan-et-al-2021}
Arkadiy Saakyan, Tuhin Chakrabarty, and Smaranda Muresan. 2021.
\newblock \href {https://doi.org/10.18653/v1/2021.acl-long.165} {{COVID}-fact:
  Fact extraction and verification of real-world claims on {COVID}-19
  pandemic}.
\newblock In \emph{Proceedings of the 59th Annual Meeting of the Association
  for Computational Linguistics and the 11th International Joint Conference on
  Natural Language Processing (Volume 1: Long Papers)}, pages 2116--2129,
  Online. Association for Computational Linguistics.

\bibitem[{Sarrouti et~al.(2021)Sarrouti, Ben~Abacha, Mrabet, and
  Demner-Fushman}]{sarrouti-et-al-2021}
Mourad Sarrouti, Asma Ben~Abacha, Yassine Mrabet, and Dina Demner-Fushman.
  2021.
\newblock \href {https://doi.org/10.18653/v1/2021.findings-emnlp.297}
  {Evidence-based fact-checking of health-related claims}.
\newblock In \emph{Findings of the Association for Computational Linguistics:
  EMNLP 2021}, pages 3499--3512, Punta Cana, Dominican Republic. Association
  for Computational Linguistics.

\bibitem[{Scepanovic et~al.(2020)Scepanovic, Martin-Lopez, Quercia, and
  Baykaner}]{scepanovic-et-at_2020}
Sanja Scepanovic, Enrique Martin-Lopez, Daniele Quercia, and Khan Baykaner.
  2020.
\newblock \href {https://doi.org/10.1145/3368555.3384467} {Extracting medical
  entities from social media}.
\newblock In \emph{Proceedings of the {ACM} Conference on Health, Inference,
  and Learning}, {CHIL} '20, pages 170--181. Association for Computing
  Machinery.
\newblock Event-place: Toronto, Ontario, Canada.

\bibitem[{Suarez-Lledo and Alvarez-Galvez(2021)}]{suarez-lledo_2021}
Victor Suarez-Lledo and Javier Alvarez-Galvez. 2021.
\newblock \href {https://doi.org/10.2196/17187} {Prevalence of health
  misinformation on social media: Systematic review}.
\newblock \emph{Journal of medical Internet research}, 23(1):e17187.

\bibitem[{Thorne and Vlachos(2018)}]{thorne-vlachos-2018}
James Thorne and Andreas Vlachos. 2018.
\newblock \href {https://aclanthology.org/C18-1283} {Automated fact checking:
  Task formulations, methods and future directions}.
\newblock In \emph{Proceedings of the 27th International Conference on
  Computational Linguistics}, pages 3346--3359, Santa Fe, New Mexico, USA.
  Association for Computational Linguistics.

\bibitem[{Thorne et~al.(2018)Thorne, Vlachos, Cocarascu, Christodoulopoulos,
  and Mittal}]{thorne-etal-2018}
James Thorne, Andreas Vlachos, Oana Cocarascu, Christos Christodoulopoulos, and
  Arpit Mittal. 2018.
\newblock \href {https://doi.org/10.18653/v1/W18-5501} {The fact extraction and
  {VER}ification ({FEVER}) shared task}.
\newblock In \emph{Proceedings of the First Workshop on Fact Extraction and
  {VER}ification ({FEVER})}, pages 1--9, Brussels, Belgium. Association for
  Computational Linguistics.

\bibitem[{Vecchi et~al.(2021)Vecchi, Falk, Jundi, and
  Lapesa}]{vecchi-etal-2021}
Eva~Maria Vecchi, Neele Falk, Iman Jundi, and Gabriella Lapesa. 2021.
\newblock \href {https://doi.org/10.18653/v1/2021.acl-long.107} {Towards
  argument mining for social good: A survey}.
\newblock In \emph{Proceedings of the 59th Annual Meeting of the Association
  for Computational Linguistics and the 11th International Joint Conference on
  Natural Language Processing (Volume 1: Long Papers)}, pages 1338--1352,
  Online. Association for Computational Linguistics.

\bibitem[{Wadden et~al.(2020)Wadden, Lin, Lo, Wang, van Zuylen, Cohan, and
  Hajishirzi}]{wadden-et-al-2020}
David Wadden, Shanchuan Lin, Kyle Lo, Lucy~Lu Wang, Madeleine van Zuylen, Arman
  Cohan, and Hannaneh Hajishirzi. 2020.
\newblock \href {https://doi.org/10.18653/v1/2020.emnlp-main.609} {Fact or
  fiction: Verifying scientific claims}.
\newblock In \emph{Proceedings of the 2020 Conference on Empirical Methods in
  Natural Language Processing (EMNLP)}, pages 7534--7550, Online. Association
  for Computational Linguistics.

\bibitem[{Wadden and Lo(2021)}]{wadden-lo-2021}
David Wadden and Kyle Lo. 2021.
\newblock \href {https://aclanthology.org/2021.sdp-1.16} {Overview and insights
  from the {SCIVER} shared task on scientific claim verification}.
\newblock In \emph{Proceedings of the Second Workshop on Scholarly Document
  Processing}, pages 124--129, Online. Association for Computational
  Linguistics.

\bibitem[{Wadden et~al.(2022)Wadden, Lo, Wang, Cohan, Beltagy, and
  Hajishirzi}]{wadden-et-al-2022}
David Wadden, Kyle Lo, Lucy Wang, Arman Cohan, Iz~Beltagy, and Hannaneh
  Hajishirzi. 2022.
\newblock \href {https://doi.org/10.18653/v1/2022.findings-naacl.6}
  {{M}ulti{V}er{S}: Improving scientific claim verification with weak
  supervision and full-document context}.
\newblock In \emph{Findings of the Association for Computational Linguistics:
  NAACL 2022}, pages 61--76, Seattle, United States. Association for
  Computational Linguistics.

\bibitem[{Wright and Augenstein(2020)}]{wright-augenstein-2020}
Dustin Wright and Isabelle Augenstein. 2020.
\newblock \href {https://doi.org/10.18653/v1/2020.findings-emnlp.43} {Claim
  check-worthiness detection as positive unlabelled learning}.
\newblock In \emph{Findings of the Association for Computational Linguistics:
  EMNLP 2020}, pages 476--488, Online. Association for Computational
  Linguistics.

\bibitem[{Wright et~al.(2022)Wright, Wadden, Lo, Kuehl, Cohan, Augenstein, and
  Wang}]{wright-et-al-2022}
Dustin Wright, David Wadden, Kyle Lo, Bailey Kuehl, Arman Cohan, Isabelle
  Augenstein, and Lucy Wang. 2022.
\newblock \href {https://doi.org/10.18653/v1/2022.acl-long.175} {Generating
  scientific claims for zero-shot scientific fact checking}.
\newblock In \emph{Proceedings of the 60th Annual Meeting of the Association
  for Computational Linguistics (Volume 1: Long Papers)}, pages 2448--2460,
  Dublin, Ireland. Association for Computational Linguistics.

\bibitem[{W{\"u}hrl and Klinger(2021)}]{wuhrl-klinger-2021}
Amelie W{\"u}hrl and Roman Klinger. 2021.
\newblock \href {https://doi.org/10.18653/v1/2021.bionlp-1.15} {Claim detection
  in biomedical {T}witter posts}.
\newblock In \emph{Proceedings of the 20th Workshop on Biomedical Language
  Processing}, pages 131--142, Online. Association for Computational
  Linguistics.

\bibitem[{Yepes and {MacKinlay}(2016)}]{yepes-macKinlay_2016}
Antonio~Jimeno Yepes and Andrew {MacKinlay}. 2016.
\newblock \href {https://www.aclweb.org/anthology/U16-1016} {{NER} for medical
  entities in twitter using sequence to sequence neural networks}.
\newblock In \emph{Proceedings of the Australasian Language Technology
  Association Workshop 2016}, pages 138--142.

\bibitem[{Yuan and Yu(2019)}]{yuan-yu-2019}
Shi Yuan and Bei Yu. 2019.
\newblock \href {https://doi.org/https://doi.org/10.1016/j.ipm.2019.03.001}
  {{H}{C}laim{E}: A tool for identifying health claims in health news
  headlines}.
\newblock \emph{Information Processing \& Management}, 56(4):1220--1233.

\bibitem[{Zhang et~al.(2021)Zhang, Li, Fukumoto, and Ye}]{zhang-etal-2021}
Zhiwei Zhang, Jiyi Li, Fumiyo Fukumoto, and Yanming Ye. 2021.
\newblock \href {https://doi.org/10.18653/v1/2021.emnlp-main.290} {Abstract,
  rationale, stance: A joint model for scientific claim verification}.
\newblock In \emph{Proceedings of the 2021 Conference on Empirical Methods in
  Natural Language Processing}, pages 3580--3586, Online and Punta Cana,
  Dominican Republic. Association for Computational Linguistics.

\bibitem[{Zuo et~al.(2020)Zuo, Acharya, and Banerjee}]{zuo-etal-2020}
Chaoyuan Zuo, Narayan Acharya, and Ritwik Banerjee. 2020.
\newblock \href {https://doi.org/10.18653/v1/2020.emnlp-main.139} {Querying
  across genres for medical claims in news}.
\newblock In \emph{Proceedings of the 2020 Conference on Empirical Methods in
  Natural Language Processing (EMNLP)}, pages 1783--1789, Online. Association
  for Computational Linguistics.

\bibitem[{Zuo et~al.(2022)Zuo, Mathur, Kela, Salek~Faramarzi, and
  Banerjee}]{zuo-et-al_2022}
Chaoyuan Zuo, Kritik Mathur, Dhruv Kela, Noushin Salek~Faramarzi, and Ritwik
  Banerjee. 2022.
\newblock \href {https://doi.org/10.1007/s41060-022-00310-7} {Beyond belief: a
  cross-genre study on perception and validation of health information online}.
\newblock \emph{International Journal of Data Science and Analytics}, pages
  1--16.

\end{thebibliography}

\clearpage
\onecolumn

\appendix

\section{Appendix}
\label{sec:appendix}

Table \ref{tab:covert-with-evidence} shows examples from the Co\textsc{Vert} dataset along with gold evidence documents and fact-checking verdicts.

\vspace{5mm}

\begingroup
    \centering \small
    \begin{tabularx}{\textwidth}{lp{54mm}Xc}
    \toprule
        
   id & full tweet & evidence & Gold \\
     \cmidrule(lr){1-1} \cmidrule(lr){2-2} \cmidrule(lr){3-3} \cmidrule(lr){4-4} 
     	
     	1a & Actually wearing masks causes bacterial pneumonia which people can die from NOT covid19. Most people do not know how to don/doff PPE properly. Follow the science Big Guy! & There's no evidence that mask-wearing causes bacterial pneumonia. & R \\
     	1b & Up to half of hospitalized COVID patients have elevated levels of antiphospholipid antibodies, or antibodies that cause blood clots to form. Patients with these antibodies are much more likely to have severe respiratory disease and kidney injury. \#COVID19 & The NIH-supported study, published in Science Translational Medicine, uncovered at least one of these autoimmune antiphospholipid (aPL) antibodies in about half of blood samples taken from 172 patients hospitalized with COVID-19. Those with higher levels of the destructive autoantibodies also had other signs of trouble. They included greater numbers of sticky, clot-promoting platelets and NETs, webs of DNA and protein that immune cells called neutrophils spew to ensnare viruses during uncontrolled infections, but which can lead to inflammation and clotting. These observations, coupled with the results of lab and mouse studies, suggest that treatments to control those autoantibodies may hold promise for preventing the cascade of events that produce clots in people with COVID-19. & S\\
     	
     	\cmidrule(lr){1-1}\cmidrule(lr){2-4}
     	
		2a & ``It’s unclear if his death was related to the virus.” This is why we perform autopsies. There is a significant likelihood that \#COVID19 played a role in that it is known to affect endothelial cells \& has been shown to cause neurological symptoms including stroke. & Some people with COVID-19 either initially have, or develop in the hospital, a dramatic state of confusion called delirium. Although rare, COVID-19 can cause seizures or major strokes. Muscular weakness, nerve injury, and pain syndromes are common in people who require intensive care during infections. & S \\

     	2b &Are you aware that the vaccines could cause miscarriage? The real data regarding covid is that there are tiny numbers, percentage wise, of generally healthy people under the age of 60 that die from COVID or that get admitted into ICU. Are you worried about cancer too? & Miscarriages have been reported following vaccination, but there's no evidence to show vaccines were the cause. The number of miscarriages reported after vaccination does not appear to exceed the number you would ordinarily expect. & R \\
     	\cmidrule(lr){1-1} \cmidrule(lr){2-4}
     	
     	3a &The predominant symptoms of `long COVID’ are psychological in nature, with anxiety and depression being most common. But those of course are also exactly the conditions which have been caused in, literally, millions of people, especially young people, by the lockdowns. & This phenomenon has led to short term as well as long term psychosocial and mental health implications for children and adolescents. The quality and magnitude of impact on minors is determined by many vulnerability factors like developmental age, educational status, pre-existing mental health condition, being economically underprivileged or being quarantined due to infection or fear of infection. & S\\
     	
     	3b & Know the facts! There is no evidence that \#COVID19 \#vaccines cause \#infertility, says @username @username \& @username \#NIAW2021 \#InfertilityAwareness & There's no evidence that approved vaccines cause fertility loss.  Although clinical trials did not study the issue, loss of fertility has not been reported among thousands of trial participants nor confirmed as an adverse event among millions who have been vaccinated. & S \\
     	
     	\cmidrule(lr){1-1} \cmidrule(lr){2-4}
		
         4a & Covid is no joke, this is why we need the vaccine. We know that mRNA doesn't cause long term effects since it decomposes in your body within 1-2 hours. Please everyone, get vaccinated as soon as you can!  
         & It's important to know that mRNA doesn't affect your genes in any way because it never enters the nucleus of cells, where your DNA is kept. After the mRNA does its job, it breaks down and is flushed out of your system within hours. &S\\
         4b & I never said Covid-19 wasn’t a real coronavirus.  And deaths linked to Covid-19 are primarily caused directly from pneumonia, or flu-like symptoms.  The classifications for influenza and pneumonia reporting changed when Covid-19 appeared. Facts. 
         & We found that septic shock and multi organ failure was the most common immediate cause of death, often due to suppurative pulmonary infection. Respiratory failure due to diffuse alveolar damage presented as immediate cause of death in fewer cases. & R \\

    \bottomrule
    \end{tabularx}
    \captionof{table}{Examples from Co\textsc{Vert} with gold evidence and fact-checking verdicts.\label{tab:covert-with-evidence}}
    \endgroup

\end{document}